%% file: acl_latex.tex
\pdfoutput=1
\documentclass[11pt]{article}

\usepackage[final]{acl}
\usepackage{changepage}

\usepackage{times}
\usepackage{latexsym}
\usepackage[T1]{fontenc}
\usepackage[utf8]{inputenc}

\usepackage{microtype}
\usepackage{inconsolata}
\usepackage{graphicx}
\usepackage{tcolorbox}
\usepackage{amsmath}
\usepackage{subcaption}
\usepackage{url} 
\usepackage{booktabs}
\usepackage{amssymb}
\usepackage{todonotes}
\usepackage{multirow}
\usepackage{colortbl}
\usepackage{xcolor}
\usepackage{makecell} 
\usepackage{arydshln}
\usepackage{fontawesome5}
\usepackage{pifont}

\usepackage{enumitem}

\usepackage{tabularx}
\usepackage{bm}

\usepackage{tikz}
\usepackage{pgfplots}
\pgfplotsset{compat=1.18}
\usetikzlibrary{arrows.meta, decorations.markings, calc, positioning, shapes.geometric}

\definecolor{landscape1}{RGB}{66, 133, 244}
\definecolor{landscape2}{RGB}{52, 168, 83}
\definecolor{trajectory}{RGB}{234, 67, 53}
\definecolor{steering}{RGB}{251, 188, 5}
\definecolor{darkgray}{RGB}{60, 60, 60}
\definecolor{boxbg}{RGB}{235, 235, 245}

\usepackage{tcolorbox}
\tcbuselibrary{skins,breakable}

\newtcolorbox{researchbox}[1]{
  colback=#1!5,
  colframe=#1!60!black,
  fonttitle=\bfseries\small,
  coltitle=black,
  enhanced,
  attach boxed title to top left={yshift=-2mm, xshift=3mm},
  boxed title style={colback=#1!30},
  sharp corners,
  breakable
}

\title{From Weights to Activations: Is Steering the Next Frontier of Adaptation?}
\newcommand{\affilsup}[1]{\rlap{\textsuperscript{\normalfont#1}}}

\author{
    Simon Ostermann\affilsup{1,2,3\textbf{*}}
    \qquad 
    Daniil Gurgurov\affilsup{1,2\textbf{*}}
    \\
    \textbf{Tanja Baeumel\affilsup{1,2,3}}
    \qquad
    \textbf{Michael A. Hedderich\affilsup{4,5}}
    \qquad
    \textbf{Sebastian Lapuschkin\affilsup{6,7}}
    \\
    \textbf{Wojciech Samek\affilsup{6,8,9}}
    \qquad
    \textbf{Vera Schmitt\affilsup{2,3,8}}
    \\
    \small{
        $^1$ Saarland University
        \quad
        $^2$ German Research Center for Artificial Intelligence (DFKI)
    }
    \\
    \small{
        $^3$ Centre for European Research in Trusted AI (CERTAIN)
        \quad
        $^4$ Center for Information and Language Processing, LMU Munich
    }
    \\
    \small{
        $^5$ Munich Center for Machine Learning
        \quad
        $^6$ Fraunhofer Heinrich Hertz Institute
        \quad
        $^7$ Technological University Dublin
    }
    \\
    \small{
        $^8$ Technische Universität Berlin
        \quad
        $^9$ Berlin Institute for the Foundations of Learning and Data (BIFOLD)
    }
    \\
    \footnotesize{\texttt{\href{mailto:simon.ostermann@dfki.de}{simon.ostermann@dfki.de}}\quad \texttt{\href{mailto:daniil.gurgurov@dfki.de}{daniil.gurgurov@dfki.de}}}
}
\begin{document}
\maketitle

\begin{abstract}

Post-training adaptation of language models is commonly achieved through parameter updates or input-based methods such as fine-tuning, parameter-efficient adaptation, and prompting. In parallel, a growing body of work modifies internal activations at inference time to influence model behavior, an approach known as \textbf{steering}. Despite increasing use, steering is rarely analyzed within the same conceptual framework as established adaptation methods.

In this work, we argue that steering should be regarded as a form of model adaptation. We introduce a set of functional criteria for adaptation methods and use them to compare steering approaches with classical alternatives. This analysis positions steering as a distinct adaptation paradigm based on targeted interventions in activation space, enabling local and reversible behavioral change without parameter updates. The resulting framing clarifies how steering relates to existing methods, motivating a unified taxonomy for model adaptation. \let\thefootnote\relax\footnotetext{*: Equal contribution. The remaining authors are sorted alphabetically.}
\end{abstract}

\section{Introduction}

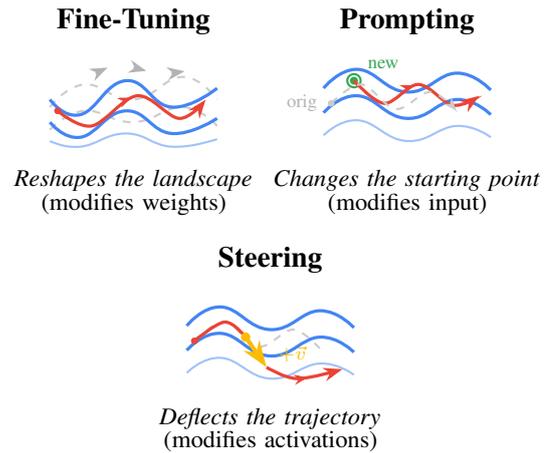
\begin{figure}[t]
\centering

\begin{tikzpicture}[scale=0.5]

\tikzset{
  every node/.style={font=\small}
}

\def\panelwidth{5.0}
\def\panelheight{2.8}
\def\spacing{2.2}

\def\vspacing{3.5}

\pgfmathsetmacro{\xshiftsteer}{\panelwidth/2 + \spacing/2}
\pgfmathsetmacro{\yshiftrow}{-(\panelheight + \vspacing)}

\begin{scope}[shift={(0,0)}]

    \node[font=\bfseries, anchor=south] at (\panelwidth/2, \panelheight+0.15) {Fine-Tuning};
    
    \draw[gray!40, dashed, thick] 
        plot[smooth, tension=0.8] coordinates {
            (0.3, 1.3) (1.2, 1.9) (2.5, 1.5) (3.8, 2.1) (4.7, 1.5)
        };
    \draw[gray!40, dashed, thick] 
        plot[smooth, tension=0.8] coordinates {
            (0.3, 0.7) (1.2, 1.2) (2.5, 0.9) (3.8, 1.3) (4.7, 0.8)
        };
    
    \draw[landscape1, very thick] 
        plot[smooth, tension=0.8] coordinates {
            (0.3, 1.5) (1.1, 1.1) (2.3, 2.0) (3.5, 1.3) (4.7, 1.8)
        };
    \draw[landscape1, very thick] 
        plot[smooth, tension=0.8] coordinates {
            (0.3, 0.9) (1.1, 0.5) (2.3, 1.2) (3.5, 0.6) (4.7, 1.0)
        };
    \draw[landscape1, thick, opacity=0.5] 
        plot[smooth, tension=0.8] coordinates {
            (0.3, 0.4) (1.1, 0.25) (2.3, 0.6) (3.5, 0.25) (4.7, 0.45)
        };
    
    \draw[trajectory, very thick, -{Stealth[length=2.5mm, width=1.8mm]},
        decoration={markings, mark=at position 0.5 with {\arrow{Stealth[length=1.8mm]}}},
        postaction={decorate}] 
        plot[smooth, tension=0.6] coordinates {
            (0.5, 1.2) (1.4, 0.8) (2.6, 1.6) (3.7, 1.0) (4.4, 1.5)
        };
    
    \fill[trajectory] (0.5, 1.2) circle (2.5pt);
    
    \foreach \x/\y/\angle in {1.5/2.2/25, 2.5/2.35/-15, 3.5/2.2/8} {
        \draw[-{Stealth}, gray!60, thick] (\x, \y) -- +(\angle:0.35);
    }
    
    \node[font=\small, text width=4.5cm, align=center, anchor=north] at (\panelwidth/2, -0.1) 
        {\textit{Reshapes the landscape}\\[-1pt]\footnotesize(modifies weights)};
\end{scope}

\begin{scope}[shift={(\panelwidth+\spacing, 0)}]
    \node[font=\bfseries, anchor=south] at (\panelwidth/2, \panelheight+0.15) {Prompting};
    
    \draw[landscape1, very thick] 
        plot[smooth, tension=0.8] coordinates {
            (0.3, 1.8) (1.2, 2.3) (2.5, 1.7) (3.8, 2.2) (4.7, 1.75)
        };
    \draw[landscape1, very thick] 
        plot[smooth, tension=0.8] coordinates {
            (0.3, 1.15) (1.2, 1.6) (2.5, 1.0) (3.8, 1.5) (4.7, 1.1)
        };
    \draw[landscape1, thick, opacity=0.5] 
        plot[smooth, tension=0.8] coordinates {
            (0.3, 0.55) (1.2, 0.95) (2.5, 0.4) (3.8, 0.8) (4.7, 0.5)
        };
    
    \fill[gray!40] (0.5, 1.4) circle (2.5pt);
    \node[gray!60, font=\scriptsize, anchor=east] at (0.4, 1.4) {orig};
    
    \fill[landscape2] (1.1, 2.0) circle (3pt);
    \draw[landscape2, thick] (1.1, 2.0) circle (5pt);
    \node[landscape2, font=\scriptsize, anchor=south west] at (1.2, 2.05) {new};
    
    \draw[trajectory, very thick, -{Stealth[length=2.5mm, width=1.8mm]},
        decoration={markings, mark=at position 0.5 with {\arrow{Stealth[length=1.8mm]}}},
        postaction={decorate}] 
        plot[smooth, tension=0.6] coordinates {
            (1.1, 2.0) (1.9, 1.4) (2.8, 1.9) (3.7, 1.35) (4.4, 1.6)
        };
    
    \draw[gray!40, dashed, thick, -{Stealth[length=1.8mm]}] 
        plot[smooth, tension=0.6] coordinates {
            (0.5, 1.4) (1.3, 1.9) (2.2, 1.25) (3.1, 1.7) (3.8, 1.2)
        };
    
    \node[font=\small, text width=4.5cm, align=center, anchor=north] at (\panelwidth/2, -0.1) 
        {\textit{Changes the starting point}\\[-1pt]\footnotesize(modifies input)};
\end{scope}

\begin{scope}[shift={({\xshiftsteer},{\yshiftrow})}]
    \node[font=\bfseries, anchor=south] at (\panelwidth/2, \panelheight+0.15) {Steering};
    
    \draw[landscape1, very thick] 
        plot[smooth, tension=0.8] coordinates {
            (0.3, 1.8) (1.2, 2.3) (2.5, 1.7) (3.8, 2.2) (4.7, 1.75)
        };
    \draw[landscape1, very thick] 
        plot[smooth, tension=0.8] coordinates {
            (0.3, 1.15) (1.2, 1.6) (2.5, 1.0) (3.8, 1.5) (4.7, 1.1)
        };
    \draw[landscape1, thick, opacity=0.5] 
        plot[smooth, tension=0.8] coordinates {
            (0.3, 0.55) (1.2, 0.95) (2.5, 0.4) (3.8, 0.8) (4.7, 0.5)
        };
    
    \fill[trajectory] (0.5, 1.4) circle (2.5pt);
    
    \draw[gray!40, dashed, thick] 
        plot[smooth, tension=0.6] coordinates {
            (0.5, 1.4) (1.3, 1.9) (2.2, 1.25) (3.1, 1.7) (3.8, 1.2)
        };
    
    \draw[trajectory, very thick] 
        plot[smooth, tension=0.6] coordinates {
            (0.5, 1.4) (1.3, 1.9) (1.85, 1.5)
        };
    
    \fill[steering] (1.85, 1.5) circle (3.5pt);
    
    \draw[steering, ultra thick, -{Stealth[length=3.5mm, width=2.5mm]}] 
        (1.85, 1.5) -- (2.4, 0.7);
    \node[steering, font=\scriptsize\bfseries, anchor=west] at (2.45, 1.05) {$+\vec{v}$};
    
    \draw[trajectory, very thick, -{Stealth[length=2.5mm, width=1.8mm]},
        decoration={markings, mark=at position 0.55 with {\arrow{Stealth[length=1.8mm]}}},
        postaction={decorate}] 
        plot[smooth, tension=0.6] coordinates {
            (2.4, 0.7) (3.2, 0.4) (4.4, 0.65)
        };
    
    \node[font=\small, text width=4.5cm, align=center, anchor=north] at (\panelwidth/2, -0.1) 
        {\textit{Deflects the trajectory}\\[-1pt]\footnotesize(modifies activations)};
\end{scope}

\end{tikzpicture}%

\caption{Conceptual illustration of three mechanisms for post-training model adaptation. Fine-tuning modifies the weight-defined behavior landscape during training, prompting alters the input-induced trajectory at inference time, and steering intervenes on internal activations during inference to deflect that trajectory.}
\label{fig:teaser}
\end{figure}

Pre-trained large language models (LLMs) form the basis of a wide range of NLP tasks, making adaptation to new tasks, domains, or behavioral constraints a central problem. Early approaches relied on full fine-tuning of models such as ELMo \cite{peters-etal-2018-deep} and BERT \cite{devlin2019bertpretrainingdeepbidirectional}, while later work introduced parameter-efficient techniques, including adapters \cite{houlsby2019parameterefficienttransferlearningnlp}, soft prompts \cite{lester2021powerscaleparameterefficientprompt}, and low-rank adaptation \cite{hu2021loralowrankadaptationlarge}. As model sizes increased, even these methods became costly, requiring substantial compute and training infrastructure; thus, producing task-specific variants became difficult to maintain at scale \cite{patterson2021carbonemissionslargeneural, bommasani2022opportunitiesrisksfoundationmodels}. There is a growing demand for adaptation methods that enable fast and flexible behavioral modification without retraining, since even parameter-efficient approaches still rely on training pipelines and hyperparameter tuning \cite{wang2025survey}. Prompting and in-context learning can only partially account for this, as both suffer from sensitivity to phrasing and example order, leading to unstable behavior \citep{chatterjee-etal-2024-posix}.

\begin{table*}[ht]
\small
\centering
\begin{tabularx}{.8\textwidth}{llXXXXXXXX}

 &
    &
  \rotatebox{45}{\faCheckCircle\ Reliability} &
  \rotatebox{45}{\faArrowRight\ Generalization} &
  \rotatebox{45}{\faStar\ Specificity} &
  \rotatebox{45}{\faThermometerHalf\ Compute Efficiency} &
  \rotatebox{45}{\faDatabase\ Data Efficiency} &
  \rotatebox{45}{\faLink\ Composability} &
  \rotatebox{45}{\faUser\ Usability} &
  \rotatebox{45}{\faUndo\ Reversibility} 
  \\ \midrule
 \parbox[t]{2mm}{\multirow{2}{*}{\rotatebox[origin=c]{90}{\tiny Prompt}}}    
                            & \textit{Prompting}                    & 0 & 0 & 0   & + & + & + & + & + \\
                            & \textit{ICL}          & 0 & 0 & 0   & + & + & + & + & + \\ \midrule
                            
 \parbox[t]{2mm}{\multirow{2}{*}{\rotatebox[origin=c]{90}{\tiny FT}}} 
                            & \textit{Fine-tuning}                   & + & + & -  & - & - & - & - & -\\
                            & \textit{RLHF}                         & + & + & -  & - & - & - & - & -\\ \midrule
                            
 \parbox[t]{2mm}{\multirow{3}{*}{\rotatebox[origin=c]{90}{\tiny PEFT}}}       
                            & \textit{Adapters}                     & + & + & 0   & + & - & + & - & +\\
                            & \textit{Soft Prompt Tuning}                 & + & + & 0   & + & 0 & + & - & +\\
                            & \textit{LoRA}                         & + & + & 0  & + & 0 & + & - & +\\ \midrule
                            
 \parbox[t]{2mm}{\multirow{3}{*}{\rotatebox[origin=c]{90}{\tiny Steering}}}                    
                            & \textit{Difference}    & + & 0 & +   & + & + & 0 & 0 & + \\
                            & \textit{Optimization}  & + & + & +   & 0 & 0 & 0 & 0 & + \\
                            & \textit{Dictionary} & 0 & + & +   & - & - & 0 & 0 & + \\ \bottomrule 
\end{tabularx}
\caption{Comparison of adaptation methods by functional criteria. \textbf{+}: a criterion is commonly demonstrated in the literature under standard settings; \textbf{-}: a criterion is systematically reported as a limitation or cost; \textbf{0}: mixed evidence, under-exploration, or high task dependence. The values for each method are justified in Sections \ref{subsec:fulfillment} and \ref{sec:evidence}.}
\label{Tab:criteria}
\end{table*}

In parallel to weight- and input-based approaches, a growing line of work has emerged from interpretability research that modifies model behavior through efficient \textbf{additive inference-time interventions on internal activations}, commonly referred to as \textit{steering}. Traditionally, such interventions were used as experimental probes to study the internal mechanisms of LLMs \cite{alain2016understanding, ivanova2021probingartificialneuralnetworks, elhage2021, marks2023geometry}. More recent work, however, demonstrates that targeted activation interventions can reliably induce new behaviors without modifying model parameters or retraining. Steering methods have been shown to influence properties such as tone, factuality, safety, and alignment  \cite{alex2023steering, panickssery2023steering, li2023inference, arditi2024refusal, konen2024stylevectorssteeringgenerative}. 

Despite increasing empirical use, steering methods are rarely analyzed within the same conceptual or evaluative framework as established adaptation techniques. Existing work primarily compares different steering approaches to one another, or to prompting baselines, with limited comparison to fine-tuning or parameter-efficient adaptation methods \cite{wuaxbench,gurgurov2026clas}. As a result, it remains unclear how steering relates to classical adaptation paradigms, which trade-offs it entails, and under which conditions it should be preferred.

In this work, we argue that steering should be viewed as a form of model adaptation rather than solely as an interpretability technique. We introduce a set of functional criteria for adaptation methods and use them to systematically compare steering approaches with fine-tuning, parameter-efficient adaptation, and prompting. Framing steering as adaptation highlights a distinct design point in which behavior is modified directly in activation space, enabling local and reversible control without parameter updates, and situates steering alongside existing adaptation paradigms. Figure~\ref{fig:teaser} illustrates the differences in common adaptation mechanisms: fine-tuning modifies the weight-defined behavior landscape, prompting alters the input-induced trajectory, and steering intervenes on internal activations during inference to deflect that trajectory. 

Our contributions are threefold:

\begin{tcolorbox}[
  enhanced,
  title=Contributions,
  fonttitle=\bfseries,
  colback=boxbg,
  colframe=black,
  coltitle=white,
  colbacktitle=black,
  boxrule=0.8pt,
  attach boxed title to top left={yshift=-2mm, xshift=8pt},
  boxed title style={colframe=black, boxrule=1pt, sharp corners},
]

\begin{itemize}[leftmargin=0.4em, itemsep=0.2em]

\item \textbf{(i)} Functional criteria for model adaptation grounded in prior work (Table~\ref{Tab:criteria}).

\item \textbf{(ii)} Systematic comparison of established adaptation methods and steering approaches under these criteria.

\item \textbf{(iii)} Conceptual argument for broadening adaptation beyond weight and input modifications to include targeted interventions on internal activations.

\end{itemize}

\end{tcolorbox}

\section{Functional Criteria for Model Adaptation}

Adaptation methods, especially for LLMs, have grown diverse. Previous work has usually concentrated on evaluating only a few isolated dimensions of adaptation, which inherently are non-exhaustive. To address this gap, we propose a set of criteria, chosen to capture the most important dimensions of adaptation. Our selection is grounded in related surveys and datasets that investigate specific adaptation methods. 

\vspace{0.2em}
\noindent\textbf{\faCheckCircle\ Reliability.}
A reliable adaptation method preserves stable behavior under repeated trials, input variation, and shifts in operating conditions within the domain. This includes consistency in quantitative metrics on data from similar domains, as well as low variance in qualitative outcomes in light of similar inputs. \citet{zhao2023survey}, for example, broadly discuss robustness and stability of LLMs, which can be subsumed under \textit{reliability}. 

\vspace{0.2em}
\noindent\textbf{\faArrowRight\ Generalization.}
Generalization reflects the capacity of the adapted model to apply learned adjustments to settings that were absent during training \cite{ben2010theory,mansour2023domainadaptationlearningbounds}. A method with strong generalization limits overfitting, maintains broad reasoning competency, and supports transfer across tasks. \citet{zhao2023survey} propose to investigate cross-task and cross-domain performance, which maps to the feature of \textit{generalization}. 
\citet{lu2025fine} survey domain adaptation techniques and their generalization potential.

\vspace{0.2em}
\noindent\textbf{\faStar\ Specificity.}
Specificity concerns the precision with which an adaptation modifies the model. High specificity yields targeted improvements in a chosen capability while reducing spillover into unrelated behaviors. This preserves the integrity of the base model and ensures controlled functional changes. For LLMs, a highly specific adaptation method results in little to no degradation of general LLM capabilities. \textit{Specificity} is included by \citet{zhao2023survey}, who review degrees of catastrophic forgetting in LLM adaptation methods. A range of work on catastrophic forgetting underlines the importance of measuring \textit{specificity} \cite{li-etal-2024-revisiting,kotha2024understanding, luo2025empirical}. 

\vspace{0.2em}
\noindent\textbf{\faThermometerHalf\ Compute Efficiency.}
Compute efficiency measures the resource demands of the adaptation process during training, along with its impact on inference. A suitable method limits training cost, memory requirements, and runtime overhead. This expands feasibility for practical applications and enables wider experimentation. \citet{lialin2023scaling} propose to evaluate PEFT methods across diverse efficiency dimensions. 

\vspace{0.2em}
\noindent\textbf{\faDatabase\ Data Efficiency.}
Data efficiency captures how well an adaptation method functions when available data is scarce or noisy. Methods with strong data efficiency extract meaningful signal from limited data and maintain performance without large training data, which is essential for specialized domains. Data efficiency is one of the most addressed dimensions in previous work 
\cite{liu2022few,pecher-etal-2025-comparing,anisuzzaman2025fine}.

\vspace{0.2em}
\noindent\textbf{\faLink\ Composability.}
Composability evaluates whether multiple adaptations can be combined without harmful interactions. A composable method supports modular updates \citet{pfeiffer-etal-2020-mad} whose effects can be analyzed and integrated in a predictable manner, simplifying iterative development and deployment across varied contexts. Composability is rarely measured directly when looking at adaptation, but a large strand of recent work tries to achieve it using parameter or activation arithmetics and other combination techniques \cite{ilharco2023editing,wang2024customizable,belanec2025task}. 

\vspace{0.2em}
\noindent\textbf{\faUser\ Usability.}
Usability reflects the ease with which an adaptation method can be applied and evaluated. This includes clarity of procedure, compatibility with established toolchains, and transparency of behavior during analysis. High usability lowers the barrier to adoption and supports reproducible research. Recent work investigates challenges non-AI experts face with prompting~\cite{10.1145/3563657.3596138,10.1145/3544548.3581388}. \citet{10.1162/tacl_a_00681} highlight brittleness issues that undermine prompting usability as an adaptation approach.

\vspace{0.2em}
\noindent\textbf{\faUndo\ Reversibility.}
Reversibility describes the extent to which an adaptation can be undone or adjusted without lasting side effects on the model. A reversible method allows behavior to be modified temporarily and incrementally, and supports rapid exploration of adaptation strength without retraining. This property is particularly important in dynamic or safety-critical settings, where adaptations may need to be enabled or disabled on the fly. Reversibility is rarely evaluated directly, as its assessment is often trivial (fine-tuning is hard to reverse; all \textit{non-invasive} adaptation techniques are easy to reverse), but a range of work on forgetting and unlearning underscores its growing importance \cite{yao2024large,geng2025comprehensive,liu2025rethinking}.

\section{Background: Classical Language Model Adaptation}

The dominant paradigm for adapting LLMs 
involves modifying the model through various training procedures. We review three major categories of classical adaptation methods that have emerged in recent years and then check their fulfillment of the functional criteria. 

\subsection{Adaptation Methods}

\vspace{0.2em}
{\textbf{Full Fine-Tuning:}}

The most straightforward approach to adapting pre-trained language models involves full fine-tuning (FFT) all model parameters on a downstream task. This approach follows the \textit{pre-train-then-finetune} paradigm established early by models pre-trained on ImageNet \cite{imagenet,Girshick_2014_CVPR,sermanet2014overfeatintegratedrecognitionlocalization} and has been adapted to NLP by models like BERT \cite{devlin2019bertpretrainingdeepbidirectional} and GPT \cite{radford2019language}, achieving strong performance across diverse tasks. FFT comes with significant computational costs and storage requirements. For instance, tuning models like Llama-3 (8-70B param-s, \citet{grattafiori2024llama}) or 
DeepSeek-V3 (671B param-s with 37B activated for each token, \citet{deepseekai2025deepseekv3technicalreport}) requires substantial GPU resources 
\cite{wan2024efficientlargelanguagemodels}. Additionally, FFT risks catastrophic forgetting \cite{french1999catastrophic} and overfitting.

Another increasingly common form of fine-tuning adapts models using feedback-driven optimization rather than supervised labels. \textit{Reinforcement Learning from Human Feedback} (RLHF) \cite{christiano2017deep, ouyang2022training} first trains a reward model from human preferences and then optimizes the base model using \textit{Proximal Policy Optimization} (PPO) \cite{schulman2017proximal} to maximize this learned reward. Newer alternatives that do not require a reward model include 
\textit{Direct Preference Optimization} (DPO, \citet{rafailov2023direct}) and \textit{Group Relative Policy Optimization} (GRPO, \citet{shao2024deepseekmath}).

\vspace{0.2em}
{\textbf{Parameter-Efficient Fine-Tuning:}}

To address efficiency challenges of FFT, Parameter-Efficient Fine-Tuning (PEFT) methods adapt pre-trained models by updating only a small subset of parameters while keeping the majority frozen \cite{ding2023parameter}. These methods substantially reduce training cost while retaining much of the performance of FFT.

\emph{Adapter layers} \cite{houlsby2019parameterefficienttransferlearningnlp, pfeiffer-etal-2020-mad} insert small trainable modules between transformer layers while freezing the base model, allowing task-specific adaptation with minimal additional parameters.

\emph{Low-Rank Adaptation (LoRA)} \cite{hu2021loralowrankadaptationlarge} assumes that adaptation-induced weight updates are low-rank, decomposing the update matrix into smaller factors. This enables training a small fraction of parameters while achieving competitive performance. Extensions such as QLoRA further reduce memory requirements \cite{dettmers2023qlora}.

\emph{Prefix tuning and prompt tuning} \cite{li-liang-2021-prefix, lester2021powerscaleparameterefficientprompt} prepend learnable vectors to inputs or intermediate representations, enabling adaptation without modifying model weights.

Despite these efficiency gains, PEFT methods still rely on gradient-based training and backpropagation. Recent surveys report over 100 PEFT variants spanning additive, reparameterized, and hybrid approaches \cite{han2024parameterefficientfinetuninglargemodels}.

\vspace{0.2em}
{\textbf{Prompting and In-Context Learning:}}

A distinct adaptation paradigm emerged with the observation that LLMs can adapt to new tasks purely through their input context, without parameter updates \cite{brown2020languagemodelsfewshotlearners}. This capability, termed \textit{in-context learning (ICL)}, enables task execution by providing natural-language instructions or demonstrations directly in the prompt \cite{dong-etal-2024-survey}. ICL spans zero-shot prompting, which relies on instruction-following behavior \cite{zhang2025instructiontuninglargelanguage}, few-shot prompting with input--output demonstrations \cite{brown2020languagemodelsfewshotlearners}, and structured strategies such as Chain-of-Thought prompting that elicit step-by-step reasoning \cite{wei2023chainofthoughtpromptingelicitsreasoning}.

ICL performance is highly sensitive to prompt design \cite{zhou2023mystery}. While ICL enables immediate, training-free adaptation, it is constrained by context length, can be less stable than fine-tuning-based approaches, and often requires careful prompt engineering \cite{dong-etal-2024-survey}. The mechanisms underlying ICL remain an active area of research 
\cite{hendel2023context}. 

\subsection{Evaluation Along Functional Criteria}
\label{subsec:fulfillment}
Table \ref{Tab:criteria} presents an overview of the fulfillment of our functional criteria for all presented methods. Detailed evidentiary support for each rating is provided in Appendix~\ref{app:evidence}.\!\footnote{Note that this functional taxonomy abstracts over performance differences, specific use cases, and implementation details; methods with similar ratings may still differ substantially in accuracy, domain suitability, or practical deployment.} 

Prompting and ICL enable adaptation without parameter updates and are highly data and compute efficient, but exhibit sensitivity to prompt phrasing and example ordering, leading to mixed reliability and generalization \citep{brown2020languagemodelsfewshotlearners,mizrahi2024state}.
FFT typically yields reliable improvements on target tasks and can generalize within the training distribution, but is computationally expensive, data intensive, and prone to catastrophic forgetting and unintended behavioral drift \citep{houlsby2019parameterefficienttransferlearningnlp}.
RLHF improves instruction following reliability and broad task performance, but requires substantial human data collection and training infrastructure and often induces global rather than targeted behavioral changes \citep{ouyang2022training}.
Adapter-based tuning introduces small task-specific modules that achieve performance close to FFT with reduced training cost, while preserving modularity and enabling composition through methods such as AdapterFusion or prompt arithmetics \citep{houlsby2019parameterefficienttransferlearningnlp,pfeiffer2021adapterfusion,belanec2025task}.
Prefix tuning and prompt tuning optimize continuous prompts with frozen model weights, offering strong data and compute efficiency and competitive generalization, particularly in low-data regimes, though with less mature usability compared to standard fine-tuning pipelines \citep{li-liang-2021-prefix,lester2021powerscaleparameterefficientprompt}.
Low-rank adaptation methods such as LoRA update a small number of parameters and match or exceed FFT performance on many tasks, while improving compute and data efficiency, though clean composition across multiple LoRA modules remains challenging \citep{hu2021loralowrankadaptationlarge,dettmers2023qlora}.

\section{Steering: Adaptation via Activation-Space Interventions}

\subsection{Early Works and Origin}

Historically, while the field has converged on weight-based modification as the standard for adaptation, the conceptual roots of influencing model behavior via internal representations are nearly as old as deep generative modeling itself. Although mechanistic interpretability has only recently emerged as a formalized discipline \cite{elhage2021, saphra2024mechanistic}, early work in Generative Adversarial Networks (GANs) \cite{goodfellow2014generative} and Variational Autoencoders (VAEs) \cite{Radford2015UnsupervisedRL} first demonstrated that latent spaces are not merely stochastic noise but possess semantically meaningful vector space arithmetic. By intervening on a latent point $z$ along a specific linear direction $v$, researchers could reliably steer output attributes, such as facial expressions, without retraining the underlying architecture \cite{Bau_2017_CVPR}. In natural language processing, this was paralleled by the discovery of the ``Sentiment Neuron'' \cite{radford2017learninggeneratereviewsdiscovering}, where fixing the activation of a single unit steered the generative process toward a desired sentiment. 

\subsection{The Linear Representation Hypothesis}

Recent theoretical and empirical work indicates that high-level concepts are represented linearly as directions in a model’s activation space \cite{elhage2021, nanda-etal-2023-emergent, park2024linearrepresentationhypothesisgeometry}. This \textit{linear representation hypothesis} (LRH) posits that concepts ranging from simple attributes such as sentiment or language identity to more complex notions like truthfulness or political ideology are encoded as vectors in internal representations, paralleling earlier findings in static word embedding spaces \cite{mikolov2013linguistic, pennington2014glove, bolukbasi2016man}.

\citet{park2024linearrepresentationhypothesisgeometry} formalize LRH using counterfactual reasoning, introducing complementary definitions in the output (word) representation space, connected to linear probing, and in the input (context) space, connected to model steering. They show that these definitions are unified through a causal inner product that preserves semantic structure and demonstrate on LLaMA-2 \cite{touvron2023llama2openfoundation} that linear representations exist for a wide range of concepts and can be used for both interpretation and control.

If concepts are encoded as directions in activation space, model behavior can be modified by intervening on activations along these directions during inference, without parameter updates. This observation underlies \textit{activation steering}, a family of methods that adapt behavior through targeted interventions on intermediate representations \cite{alex2023steering, zou2023representation, li2023inference}.

The view of steering as adaptation is further supported by the framework of causal abstraction built on top of the LRH \cite{NEURIPS2021_4f5c422f, NEURIPS2022_6f1d43d5, geiger2024causal}. This framework characterizes how high-level causal models can faithfully abstract neural networks and provides a theoretical basis for intervention-based adaptation. Steering methods satisfy the criteria of causal interventions: they target specific causal variables, modify them in controlled ways, and induce predictable downstream effects. This causal framing also connects steering to mechanistic interpretability methods such as activation patching \cite{meng2022, meng_locating_2023} and causal tracing \cite{genderbias}, which employ similar interventions for analysis rather than adaptation.

It is worth noting that our argument does not require that the LRH holds universally across all concepts, layers, or models. Recent work has shown that under relaxed linearity constraints, some representational alignment claims become vacuous \citep{sutter2025non}. We rely here on the weaker claim that approximate linear structure is sufficiently stable and widespread to support reliable intervention in practice. Empirical evidence demonstrates that steering works robustly across diverse concepts, models, and tasks \citep{alex2023steering, arditi2024refusal, rimsky-etal-2024-steering}, suggesting that even if linearity is approximate or local, it provides a useful basis for adaptation. The success of steering methods does not depend on perfect linearity, but on the existence of intervention points where additive or low-rank modifications yield predictable behavioral effects.

\subsection{Steering Methods}

Activation steering modifies model behavior by intervening on intermediate activations during the forward pass. The general approach involves three steps: (1) identifying a concept of interest, (2) computing a steering vector that captures this concept's direction in activation space, and (3) adding this vector to the model's activations at specific layers during generation.

{\textbf{Difference-Based Steering:}}

\textit{Activation Addition} (ActAdd) \cite{alex2023steering} computes steering vectors by taking the difference in activations between pairs of contrasting prompts (e.g., "Love" vs. "Hate", "Truthful" vs. "Deceptive"). \textit{Contrastive Activation Addition} (CAA) \cite{rimsky-etal-2024-steering} extends this approach by averaging activation differences across multiple positive and negative examples of a behavior, producing more robust steering vectors. \textit{Difference-in-Means} (DiffMean) \cite{marks2023geometry} provides theoretical grounding for these approaches, showing that the difference between mean activations of two classes captures a causally relevant direction for steering.

{\textbf{Optimization-based Steering:}}

Optimization-based approaches learn steering interventions through supervised optimization in activation space. \textit{Representation Fine-Tuning} (Reft) \cite{wu2024reft} learns task-specific interventions by optimizing low-rank updates to hidden representations. The method introduces learnable intervention functions that modify activations at selected layers and positions, achieving parameter efficiency comparable to LoRA. Building on this, \textit{Reft-r1} \cite{wuaxbench} constrains the interventions to rank-1 subspaces, jointly learning concept detection and steering through a unified objective. 

{\textbf{Dictionary Learning Steering:}}

\textit{Sparse Autoencoders} (SAEs) \cite{gao2024scaling, templeton_scaling_2024, huben2024sparse} represent a self-supervised approach to discovering steering directions. By learning sparse representations of model activations, SAEs decompose the activation space into interpretable features that can be individually manipulated for steering. While SAEs have shown promise for interpretability, \citet{wuaxbench}'s evaluation shows that they underperform DiffMean- and optimization-based methods for steering tasks. \citet{arad-etal-2025-saes} improves steering with SAEs through smarter feature selection. 

\begin{figure}[ht!]
\centering
\begin{tikzpicture}
\small
  \begin{axis}[
    width=0.9\linewidth,
    height=5cm,
    xlabel=Year,
    ylabel=Proportion (\%),
    legend columns=4,
    legend style={at={(0.5,1.08)}, anchor=south, font=\small},
    grid=major,
    grid style={gray!30},
    xtick={2020,2021,2022,2023,2024,2025},
    xticklabels={20,21,22,23,24,25},
    ymajorgrids=true,
    xmajorgrids=false,
    ymax=85,
  ]
  
  \definecolor{cbblack}{HTML}{000000}
  \definecolor{cborange}{HTML}{E69F00}
  \definecolor{cbblue}{HTML}{56B4E9}
  \definecolor{cbgreen}{HTML}{009E73}
  
  \addplot[line width=2pt, mark=o, color=cbblack] coordinates {
    (2020, 6.4) (2021, 12.8) (2022, 10.7) (2023, 8.9) (2024, 8.8) (2025, 8.5)
  };
  \addlegendentry{PEFT}
  
  \addplot[line width=2pt, mark=square, color=cborange] coordinates {
    (2020, 82.7) (2021, 68.5) (2022, 38.6) (2023, 27.6) (2024, 26.1) (2025, 22.8)
  };
  \addlegendentry{FFT}
  
  \addplot[line width=2pt, mark=triangle, color=cbblue] coordinates {
    (2020, 8.2) (2021, 16.9) (2022, 47.6) (2023, 61.2) (2024, 61.7) (2025, 62.1)
  };
  \addlegendentry{Prompt.}
  
  \addplot[line width=2.5pt, mark=diamond, color=cbgreen] coordinates {
    (2020, 2.7) (2021, 1.8) (2022, 3.0) (2023, 2.2) (2024, 3.3) (2025, 6.5)
  };
  \addlegendentry{Steer.}
  
  \end{axis}
\end{tikzpicture}
\caption{Relative share of adaptation techniques across *CL and NeurIPS conference abstracts (2020-2025). Steering's proportion grows, while fine-tuning's dominance declines and prompting becomes the focus.}
\label{fig:adaptation_proportions_alt1}
\end{figure}
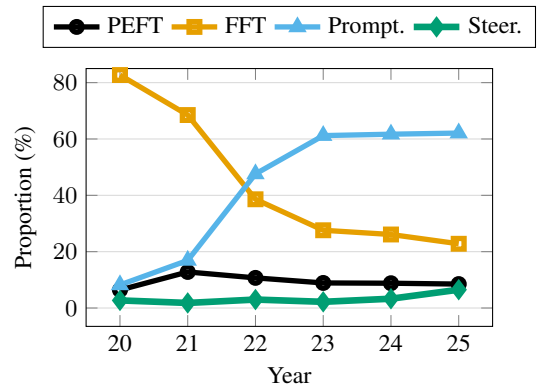

\section{Empirical Evidence of Steering as Adaptation}
\label{sec:evidence}

Having established the functional criteria and theoretical foundations of steering, we examine empirical evidence showing that steering can function as an adaptation method across application domains. Figure~\ref{fig:adaptation_mentions_alt1}\footnote{Details on the crawling procedure are in Appendix \ref{app:crawling}.} contextualizes this analysis by illustrating the growing prevalence of steering relative to other adaptation techniques in *CL and NeurIPS conference abstracts, while the following sections justify the assessments summarized in Table~\ref{Tab:criteria}. Overall, steering occupies a distinct region of the adaptation space characterized by high specificity, efficiency, and reversibility, though often with weaker global guarantees, and its usability remains under-explored, precluding a structured evaluation.

\subsection{Instruction Following and General Control}

Several lines of work demonstrate that steering can improve one of the most basic skills of language models, namely instruction following.

\citet{liu2024incontext} demonstrate that in-context vectors, latent-state shifts computed from few-shot examples, can steer models to perform new tasks without retraining or additional prompt tokens, achieving performance comparable to or better than standard ICL, i.e. more \textit{reliably}, while being substantially more \textit{compute efficient}. \citet{todd2024functionvectorslargelanguage} identify function vectors in specific attention heads that encode task-specific input-output mappings; inserting these vectors into unrelated contexts reliably triggers the corresponding task, indicating \textit{generalization} beyond the training distribution.

For text generation control, \citet{alex2023steering} show that steering vectors computed from contrastive prompt pairs can control sentiment, topic, and writing style, achieving strong performance while requiring only paired prompts rather than labelled data, demonstrating high \textit{data efficiency}. \citet{konen2024stylevectorssteeringgenerative} extend this approach to fine-grained stylistic attributes such as emotional tone, formality, and authorial voice. For \textit{composability}, \citet{ilharco2022editing} show that task vectors can be additively combined to control multiple attributes simultaneously, while \citet{subramani2022extracting} provide early evidence that latent steering vectors exhibit vector arithmetic properties for sentiment and topic control.

\subsection{Safety, Alignment, and Transfer}

Inference-time activation interventions can reliably modify alignment-relevant behaviors without retraining, including increasing model truthfulness through attention head interventions while preserving general benchmark performance \citep{li2023inference, zou2023representation}, and adapting intervention strength per context to further improve truthfulness without degrading task accuracy \citep{bayat2024enhanced}. Similarly, residual stream interventions can increase or decrease sycophancy with effectiveness comparable to fine-tuning while preserving general knowledge performance \citep{panickssery2023steering}, illustrating that steering enables behavioral modification without parameter updates and satisfies \textit{reliability}, \textit{compute efficiency}, and \textit{specificity}.

Safety-critical behaviors such as refusal can be mediated by low-dimensional activation subspaces that can be reliably controlled through minimally invasive steering interventions, with effects generalizing across prompts, languages, and large test sets from very few examples, and even reproducing behaviors induced by reinforcement learning \citep{arditi2024refusal, rimsky-etal-2024-steering, wang2025refusal,sinii2025small}, demonstrating strong \textit{generalization} and \textit{data efficiency}.

\subsection{Multilinguality}

Activation steering has been applied to multilingual and cross-lingual adaptation by identifying and manipulating language-specific activation features to improve cross-lingual performance without retraining \citep{zhao2024large, tang2024language, gurgurov-etal-2025-language, gurgurov2026clas}, enable controlled language switching while preserving semantics \citep{chou2025causallanguagecontrolmultilingual}, and selectively impact individual languages through targeted SAE feature interventions, demonstrating \textit{zero-shot domain adaptation} and \textit{specificity} \citep{deng-etal-2025-unveiling}.

Steering has been extended to vision-language models through modality-specific inference-time interventions for control, safety, and hallucination mitigation \citep{wang2024inferaligner, sivakumar2025steervlm, li2025hidden, su2025activation}, suggesting that steering as an activation-level adaptation paradigm \textit{generalizes} beyond text-only settings.

\section{Conceptual Argument}
The evidence presented in Section \ref{sec:evidence} demonstrates how steering satisfies the functional criteria for adaptation. In this section, we argue that recognizing steering as adaptation is not merely terminological but a reframing with practical consequences. 

\subsection{Challenging the Weight-Modification Assumption}

Classical adaptation paradigms--FFT, PEFT, and even prompting--share an implicit assumption about what constitutes adaptation. Fine-tuning and PEFT assume adaptation requires weight modification, while prompting assumes adaptation occurs through input manipulation. Both perspectives treat the model's internal activations as consequences of adaptation rather than as a site of adaptation itself.

The LRH challenges this assumption. If concepts are encoded as directions in activation space, and if behavior is determined by trajectories through that space, then modifying those trajectories directly constitutes a legitimate form of behavioral change. Weights define a \textit{potential} landscape of behaviors; activations determine which behaviors are actually realized. In simple words: classical fine-tuning reshapes the landscape while steering modifies the path taken through that landscape (s. Figure \ref{fig:teaser}). Both achieve behavioral change, but through different mechanisms.

These observations suggest a broader, \textit{functional} definition: \textbf{Adaptation is any systematic method that reliably modifies model behavior to meet new requirements}. Under this definition, the mechanism, whether weight updates, input manipulation, or activation intervention, is secondary to the functional outcome.

\subsection{Why the Reframing Matters}

Recognizing steering as adaptation is not merely semantic; it expands the design space for post-training control by introducing an additional axis of behavioral modification. \textbf{Efficiency and reversibility} arise from the fact that steering interventions can be applied, removed, or adjusted instantly, enabling dynamic control without retraining. \textbf{Interpretability} follows because steering operates on directions that have the potential to correspond to human-interpretable concepts, keeping adaptation targets explicit and allowing changes to be traced to specific internal components such as neurons, heads, layers, or subspaces. \textbf{Knowledge preservation} is achieved insofar as steering can instill desired behaviors without broadly affecting general capabilities, mitigating catastrophic forgetting commonly associated with fine-tuning. Finally, \textbf{practical deployment} is enabled by the absence of gradient computation, allowing steering to be used for bias correction, safety enforcement, or contextual customization without the infrastructure required for retraining.

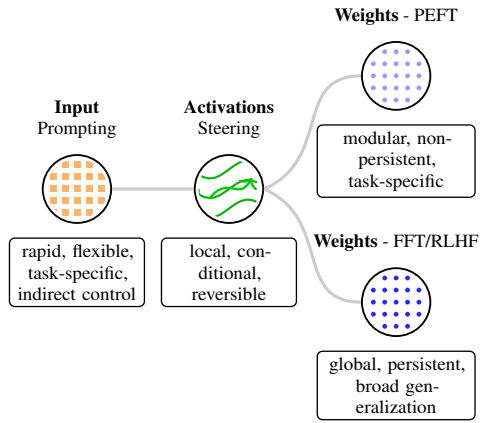
\begin{figure}[t]
\centering
\begin{tikzpicture}[
    font=\small,
    knob/.style={draw, circle, minimum size=0.9cm, line width=0.7pt},
    label/.style={align=center, font=\scriptsize},
    desc/.style={align=center, font=\scriptsize, draw, rounded corners=2pt, inner sep=2.5pt, line width=0.5pt}
]

\node[knob] (prompt) at (0,0) {};
\begin{scope}
    \clip (prompt) circle (0.4cm);
    \foreach \x in {-0.36,-0.2,-0.04,0.12,0.28} {
        \foreach \y in {-0.36,-0.2,-0.04,0.12,0.28} {
            \fill[orange!60] ($(prompt.center)+(\x,\y)$) rectangle ++(0.09,0.09);
        }
    }
\end{scope}
\node[label, above=0.08cm of prompt] {\textbf{Input}\\Prompting};
\node[desc, below=0.18cm of prompt, text width=1.6cm] {rapid, flexible,\\task-specific,\\indirect control};

\node[knob] (act) at (2.0,0) {};
\begin{scope}
    \clip (act) circle (0.4cm);
    \foreach \y in {-0.26, 0, 0.26} {
        \draw[green!70!black, thick]
        ($(act.center)+(-0.4,\y)$)
        \foreach \x in {-0.4,-0.33,...,0.4} {
            -- ($(act.center)+(\x,{0.12*sin(360*\x/1.0)+\y})$)
        };
    }
    \draw[green!80!black, ->, line width=0.9pt]
        ($(act.center)+(-0.26,-0.13)$) .. controls ($(act.center)+(0,0.2)$) and ($(act.center)+(0.13,-0.18)$) .. ($(act.center)+(0.3,0.15)$);
\end{scope}
\node[label, above=0.08cm of act] {\textbf{Activations}\\Steering};
\node[desc, below=0.18cm of act, text width=1.6cm] {local, conditional,\\reversible};

\node[knob] (peft) at (4.2,1.5) {};
\begin{scope}
    \clip (peft) circle (0.4cm);
    \foreach \x in {-0.3,-0.15,0,0.15,0.3} {
        \foreach \y in {-0.32,-0.16,0,0.16,0.32} {
            \fill[blue!40] ($(peft.center)+(\x,\y)$) circle (0.032);
        }
    }
\end{scope}
\node[label, above=0.08cm of peft] {\textbf{Weights} - PEFT};
\node[desc, below=0.18cm of peft, text width=1.9cm] {modular, non-persistent,\\task-specific};

\node[knob] (ft) at (4.2,-1.5) {};
\begin{scope}
    \clip (ft) circle (0.4cm);
    \foreach \x in {-0.3,-0.15,0,0.15,0.3} {
        \foreach \y in {-0.32,-0.16,0,0.16,0.32} {
            \fill[blue!85] ($(ft.center)+(\x,\y)$) circle (0.032);
        }
    }
\end{scope}
\node[label, above=0.08cm of ft] {\textbf{Weights} - FFT/RLHF};
\node[desc, below=0.18cm of ft, text width=1.9cm] {global, persistent,\\broad generalization};

\draw[gray!40, line width=1.2pt] (prompt.east) -- (act.west);

\draw[gray!40, line width=1.2pt] (act.east) .. controls ++(0.9,0.4) and ++(-0.9,0) .. (peft.west);
\draw[gray!40, line width=1.2pt] (act.east) .. controls ++(0.9,-0.4) and ++(-0.9,0) .. (ft.west);
\end{tikzpicture}
\caption{Choosing the right adaptation method.}
\label{fig:whenwhat}
\end{figure}

\subsection{Which Adaptation Method to Choose?}
The reframing of steering as adaptation clarifies when different forms of adaptation are appropriate by highlighting the level at which behavior is modified (s. Figure \ref{fig:whenwhat}). \textbf{Weight-based methods}, such as FFT and RLHF, redefine the model’s overall behavior landscape and are therefore most suitable when the desired change is \textbf{global, persistent, and expected to generalize broadly} across inputs. PEFT complements this with \textbf{non-persistent, modular, and more task-specific} methods.

\textbf{Input-based adaptation through prompting} operates at a different level. Prompting preserves the underlying model and instead influences behavior by shaping the context in which the model operates. This makes it well suited for \textbf{rapid, task-specific, and exploratory adaptation}, where flexibility and zero training cost are priorities. At the same time, prompting relies on indirect control through natural language instructions and demonstrations, which can lead to sensitivity to phrasing, ordering, and context length, limiting reliability and specificity in some settings.

In contrast, \textbf{activation-level adaptation} modifies behavior locally along specific internal trajectories during inference. This makes steering particularly meaningful when changes are \textbf{contextual, conditional, or temporary}, and when preserving the original model behavior and parameters is important. From this perspective, the choice between weight-based, input-based, and activation-based adaptation reflects different assumptions about the scope, stability, and reversibility of the intended change, rather than being a matter of efficiency alone.

\section{Challenges and Future Directions}

\subsection{Open Challenges}

While activation steering enables efficient and targeted behavior modification, it does not by itself provide guarantees about robustness, safety, or long-term stability. Steering should therefore not be viewed as a complete solution to model control, but as a modular mechanism that must be embedded within broader systems of verification, monitoring, and constraint enforcement. Going forward, establishing steering as a full alternative technique for model adaptation, several challenges arise.

\noindent\textbf{$\bm{{(i)}}$ Entangled representations and side effects.} Because model representations often encode multiple concepts in superposition, steering interventions targeting one attribute may inadvertently affect others \cite{siu2025steeringsafety}. Non-orthogonal steering directions can interfere with each other, and modifications intended for specific behaviors may propagate to unrelated capabilities through entangled internal representations \cite{raedler2025the}. This necessitates thorough evaluation of off-target effects and careful selection of intervention points.

\noindent\textbf{$\bm{{(ii)}}$ Interaction with alignment mechanisms.} Steering may interact with behaviors instilled through FFT or RLHF. These interactions are not yet well-understood, and steering interventions could potentially circumvent or conflict with the alignment properties acquired during training \cite{raedler2025the}. Understanding how steering composes with other adaptation methods, particularly safety-critical alignment procedures, remains an important open question \cite{stickland2024steering}.

\noindent\textbf{$\bm{{(iii)}}$ Modular control with guarantees.} A key direction for future work is the development of modular control architectures in which steering interventions operate as explicitly defined components with well understood interfaces and failure modes \cite{stickland2024steering,postmus2024steering,wang2025expertsteer}. Such systems would combine steering with complementary mechanisms such as behavioral verification or runtime monitoring,  enabling stronger guarantees than any single technique can provide in isolation. In this view, steering supplies flexible local control, while other components enforce global constraints.

\noindent\textbf{$\bm{{(iv)}}$ Compositionality beyond vector addition.} Although many steering methods exploit linear structure that supports additive composition, reliable compositionality remains an open challenge. Interference between non orthogonal concepts, conflicts between objectives, and sensitivity to intervention scale limit current approaches. Addressing these issues requires moving beyond purely algebraic composition toward structured representations of goals and constraints \cite{nguyen-etal-2025-multi}.

\noindent\textbf{$\bm{{(v)}}$ Steering and interpretability by design.} The effectiveness of steering depends on the presence of semantically meaningful structure in a model’s internal representations, which suggests a deeper connection between steering and long-term goals in model design. Rather than treating steering as a post-hoc technique, future architectures may be explicitly designed to expose interpretable, modular control points that support reliable intervention \cite{gao2025weight}. Steering can inform the development of models that are controllable and interpretable by design.

\subsection{Towards Steering as Standard Practice for Adaptation}

Treating steering as adaptation requires concrete shifts in research and deployment practices. We identify four priorities:

\noindent\textbf{$\bm{{(i)}}$ Shared evaluation standards.} Steering methods need systematic comparison to prompting, fine-tuning, and PEFT, using standardized benchmarks. Evaluation should cover task accuracy, generalization to held-out domains, compositional behavior under conflicting objectives, and robustness to distribution shifts. This requires benchmark suites spanning diverse tasks and models, with metrics for both target effects and unintended side effects.

\noindent\textbf{$\bm{{(ii)}}$ Mature tooling.} Steering needs infrastructure comparable to Hugging Face PEFT \cite{peft}: pre-computed vector libraries for common tasks, validated layer selection recipes, and compositional frameworks for combining interventions. Current practice relies on custom implementations and manual tuning. 

\noindent\textbf{$\bm{{(iii)}}$ Design-level integration.} Researchers and developers should evaluate whether behavioral changes are better achieved through weights, inputs, or activations based on persistence, scope, usability, and interpretability requirements. Steering should be a first-class design option alongside FFT and prompting, not an experimental add-on.

\noindent\textbf{$\bm{{(iv)}}$ Documentation and pedagogy.} Developers should investigate steering compatibility alongside fine-tuning benchmarks, indicating which architectures support interpretable intervention and at which layers. Steering should be mentioned in core NLP curricula as a standard adaptation technique. Practitioners need decision frameworks clarifying when steering outperforms alternatives.

\section{Conclusion}

This work argues that steering constitutes a genuine form of model adaptation, distinct from weight- and input-based approaches. By intervening directly in activation space, steering enables local, efficient, and reversible behavioral modification. 
Framing steering as adaptation clarifies its relationship to FFT, PEFT, and prompting and highlights a previously under-articulated design point in the space of post-training control mechanisms.

Treating steering as adaptation expands the set of tools available for modifying model behavior, particularly in settings where changes must be targeted, temporary, or interpretable. Future work can build on this framing by developing more systematic evaluations of steering methods, exploring their interaction with other adaptation techniques, and investigating how architectural choices affect the structure and controllability of activation space.

\section*{Limitations}

This work is not intended as a systematic or exhaustive literature review of steering methods or adaptation techniques. Our goal is conceptual rather than taxonomic: we focus on articulating an argument for viewing steering as a form of model adaptation and on positioning it relative to established paradigms using a small set of functional criteria.

As a consequence, our coverage of steering methods is necessarily selective. While we draw on representative examples spanning difference-based, optimization-based, and dictionary-based approaches, we do not claim to cover all existing or emerging variants of steering, nor do we provide a comprehensive empirical comparison across methods. Some recent or highly specialized approaches may therefore fall outside the scope of our discussion.

Additionally, our evaluation of adaptation methods relies on a qualitative synthesis of results reported in prior work rather than on new large-scale experimental benchmarks. Although this reflects common practice for conceptual and position papers, it means that the assessments summarized in Table~\ref{Tab:criteria} should be interpreted as \textit{indicative rather than definitive}.

We view these limitations as a trade-off in service of clarity and focus. A systematic survey or benchmark-driven comparison of steering methods would be a valuable direction for future work, but lies beyond the scope of the present argument-driven analysis.

Finally, our conceptual framing and many steering techniques implicitly rely on assumptions about how concepts and behaviors are encoded in model representations, such as the idea that meaningful directions or mappings are approximately linear; recent work has shown that when one lifts the linearity constraint in representation mappings, many model-to-algorithm alignment methods become vacuous, underscoring that linear representational structure is a modeling assumption that may not hold uniformly across tasks, layers, or models \citep{sutter2025non}.

Viewing steering as adaptation and explainability raises several risks. Activation-level interventions can give a misleading sense of understanding or control, as modifying behavior does not imply that the underlying reasoning is fully captured, causally proven, or faithfully explained. Because steering operates at inference time without parameter updates, it can be applied dynamically, complicating attribution and accountability in deployed systems. In addition, steering directions may interact in unintended ways, producing side effects on unrelated behaviors, particularly when internal representations are entangled. Finally, steering relies on assumptions about representational structure, such as approximate linearity and stability, which may not hold uniformly across models or settings. These considerations highlight that steering should be used cautiously and as part of broader evaluation and oversight mechanisms. We emphasize that this work does not propose steering as a complete solution to model explainability or control, but rather as a conceptual lens that clarifies where and how behavior can be modified. Recognizing these risks is essential for the responsible use and future development of activation-based adaptation methods.

\section*{Acknowledgments}
AI assistance was used to improve the clarity and fluency of the writing, to help refine phrasing and structure, and to support exploratory literature search and organization. All scientific claims, interpretations, and conclusions remain the responsibility of the authors. This work was supported by the German Federal Ministry of Research, Technology and Space (BMFTR) as part of the project TRAILS (01IW24005).
\bibliography{custom}

\clearpage

\appendix
\newpage

\section*{Appendix}
\input{Appendices/CrawlingDescription}

\input{Appendices/Evidence}

\end{document}

%% file: Appendices/CrawlingDescription.tex
\section{Automatic Identification of Adaptation Papers in the ACL Anthology and NeurIPS proceedings}
\label{app:crawling}
We automatically estimated the number of adaptation papers per year using the ACL Anthology and NeurIPS proceedings. For each venue: ACL, NAACL, EMNLP, EACL, and Findings from 2020 to 2025, we crawled all conference volumes and extracted paper titles and abstracts from their landing pages. Papers were assigned to adaptation categories based on simple keyword matching in the abstract: PEFT using “adapter”, “lora”, or “peft”; finetuning using “fine-tune” or “finetune”; prompting using “prompt”; and steering using “steer”. We selected relevant NeurIPS papers from 2020 to 2025 in the same way, with the added constraint that either of the keywords “LLM” or “language” appeared in the abstract to focus on the adaptation of language models. 
Table \ref{tab:adaptation_counts} shows the total number of relevant papers per year.

To compute yearly counts and proportions, we deduplicated papers across categories and venues so that each paper contributed at most once per year to the total denominator while still counting toward all categories it matched. Relative shares were computed by normalizing category counts by the total number of unique adaptation papers per year.

\begin{table}[ht]
\centering
\setlength{\tabcolsep}{3pt}
\begin{tabular}{@{}c|cccc@{}}
\hline
Year & PEFT & Finetuning & Prompting & Steering \\
\hline
\multicolumn{5}{c}{\textbf{CL Conferences}} \\
\hline
2020 & 6 & 83 & 9 & 3 \\
2021 & 27 & 146 & 32 & 4 \\
2022 & 56 & 187 & 236 & 15 \\
2023 & 94 & 303 & 632 & 22 \\
2024 & 135 & 428 & 1045 & 57 \\
2025 & 166 & 543 & 1445 & 128 \\
\hline
\multicolumn{5}{c}{\textbf{NeurIPS}} \\
\hline
2020 & 1 & 8 & 0 & 0 \\
2021 & 1 & 4 & 5 & 0 \\
2022 & 4 & 29 & 30 & 2 \\
2023 & 17 & 42 & 132 & 6 \\
2024 & 52 & 127 & 268 & 14 \\
2025 & 80 & 116 & 350 & 60 \\
\hline
\end{tabular}
\caption{Number of adaptation papers per year by topic for CL conferences and NeurIPS.}
\label{tab:adaptation_counts}
\end{table}

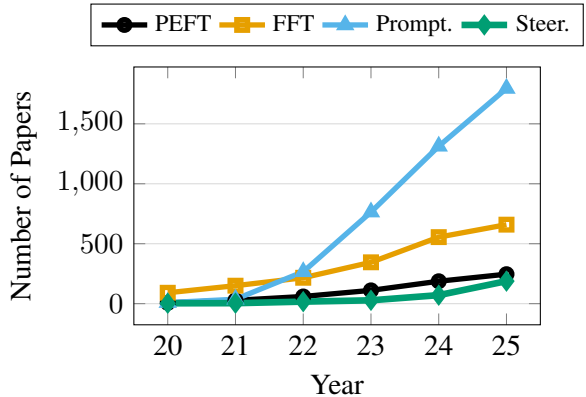
\begin{figure}[t]
\centering
\begin{tikzpicture}
  \begin{axis}[
    width=0.9\linewidth,
    height=5cm,
    xlabel=Year,
    ylabel=Number of Papers,
    legend columns=4,
    legend style={at={(0.5,1.08)}, anchor=south, font=\small},
    grid=major,
    grid style={gray!30},
    xtick={2020,2021,2022,2023,2024,2025},
    xticklabels={20,21,22,23,24,25},
    ymajorgrids=true,
    xmajorgrids=false,
  ]

  \definecolor{cbblack}{HTML}{000000}
  \definecolor{cborange}{HTML}{E69F00}
  \definecolor{cbblue}{HTML}{56B4E9}
  \definecolor{cbgreen}{HTML}{009E73}
  
  \addplot[line width=2pt, mark=o, color=cbblack] coordinates {
    (2020, 7) (2021, 28) (2022, 60) (2023, 111) (2024, 187) (2025, 246)
  };
  \addlegendentry{PEFT}
  
  \addplot[line width=2pt, mark=square, color=cborange] coordinates {
    (2020, 91) (2021, 150) (2022, 216) (2023, 345) (2024, 555) (2025, 659)
  };
  \addlegendentry{FFT}
  
  \addplot[line width=2pt, mark=triangle, color=cbblue] coordinates {
    (2020, 9) (2021, 37) (2022, 266) (2023, 764) (2024, 1313) (2025, 1795)
  };
  \addlegendentry{Prompt.}
  
  \addplot[line width=2.5pt, mark=diamond, color=cbgreen] coordinates {
    (2020, 3) (2021, 4) (2022, 17) (2023, 28) (2024, 71) (2025, 188)
  };
  \addlegendentry{Steer.}
  
  \end{axis}
\end{tikzpicture}
\caption{Mentions of adaptation techniques across *CL and NeurIPS conference abstracts (2020--2025). Steering shows accelerating growth from 3 papers in 2020 to 188 in 2025, while prompting dominates and fine-tuning's share declines.}
\label{fig:adaptation_mentions_alt1}
\end{figure}

%% file: Appendices/Evidence.tex
\section{Evidentiary Support for Table 1}
\label{app:evidence}
 
This appendix provides the evidentiary basis for the qualitative assessments in Table \ref{Tab:criteria}. For each adaptation method, we summarize the empirical findings that support our ratings across the eight functional criteria. These assessments reflect synthesis across the cited literature rather than exhaustive coverage, and individual implementations may deviate from these characterizations.

\clearpage
\subsection{Prompting and In-Context Learning}

\vspace{0.5em}
\noindent
\begin{minipage}{\textwidth}
\centering
\begin{tabularx}{\textwidth}{l c X}
\toprule
\textbf{Criterion} & \textbf{Rating} & \textbf{Evidence} \\
\midrule
\faCheckCircle Reliability & 0 & Prompting exhibits high sensitivity to phrasing, example ordering, and formatting \citep{10.1162/tacl_a_00681}, leading to inconsistent performance across seemingly equivalent formulations. \\
\addlinespace
\faArrowRight Generalization & 0 & Transfer depends heavily on task similarity and prompt design \citep{zhou2023mystery}, with mixed results across domains. \\
\addlinespace
\faStar Specificity & 0 & Prompting influences model behavior globally through the input context. Subtle prompt changes intended to improve one aspect can have unintended effects on unrelated behaviors, a phenomenon known as ``prompt drift'' \citep{commey2026better}. \\
\addlinespace
\faThermometerHalf Compute Efficiency & + & No training required; adaptation occurs at inference time with minimal overhead \citep{brown2020languagemodelsfewshotlearners}. \\
\addlinespace
\faDatabase Data Efficiency & + & Few-shot and zero-shot settings demonstrate strong performance with minimal examples \citep{brown2020languagemodelsfewshotlearners}. \\
\addlinespace
\faLink Composability & + & Multiple instructions and demonstrations can be naturally combined in the prompt context. \\
\addlinespace
\faUser Usability & + & The natural language interface makes prompting accessible without specialized ML expertise. \\
\addlinespace
\faUndo Reversibility & + & Behavioral changes are ephemeral and tied to the prompt; no persistent model modification occurs. \\
\bottomrule
\end{tabularx}
\captionof{table}{Evidentiary support for Prompting and In-Context Learning ratings in Table \ref{Tab:criteria}.}
\label{tab:evidence_prompting}
\end{minipage}

\clearpage
\subsection{Full Fine-Tuning and RLHF}

\vspace{0.5em}
\noindent
\begin{minipage}{\textwidth}
\centering
\begin{tabularx}{\textwidth}{l c X}
\toprule
\textbf{Criterion} & \textbf{Rating} & \textbf{Evidence} \\
\midrule
\faCheckCircle Reliability & + & Fine-tuning achieves stable, consistent performance on target tasks \citep{devlin2019bertpretrainingdeepbidirectional}. RLHF reliably improves instruction following and alignment \citep{ouyang2022training}. \\
\addlinespace
\faArrowRight Generalization & + & Strong transfer within the training distribution and to similar domains is trivially exhibited by any modern LLM with zero-shot capabilities, usually acquired via fine-tuning and RLHF during post-training. \\
\addlinespace
\faStar Specificity & $-$ & Both methods induce global behavioral changes and are prone to catastrophic forgetting of general capabilities \citep{french1999catastrophic, luo2025empirical}. \\
\addlinespace
\faThermometerHalf Compute Efficiency & $-$ & High computational cost for training, especially for large models \citep{patterson2021carbonemissionslargeneural}. RLHF adds further overhead from reinforcement learning. \\
\addlinespace
\faDatabase Data Efficiency & $-$ & Training modern LLMs requires substantial training data; RLHF additionally requires extensive human preference annotations. \\
\addlinespace
\faLink Composability & $-$ & Multiple fine-tuning objectives interfere; combining adapted models is non-trivial. Sequential training of SFT and RLHF leads to catastrophic forgetting \citep{fernando2024understanding}. \\
\addlinespace
\faUser Usability & $-$ & Requires significant infrastructure, expertise, and hyperparameter tuning. RLHF pipelines are particularly complex. \\
\addlinespace
\faUndo Reversibility & $-$ & Model weights are permanently modified; reverting requires retraining from the base model. \\
\bottomrule
\end{tabularx}
\captionof{table}{Evidentiary support for Full Fine-Tuning and RLHF ratings in Table \ref{Tab:criteria}.}
\label{tab:evidence_finetuning}
\end{minipage}

\clearpage
\subsection{Parameter-Efficient Fine-Tuning}

\subsubsection{Adapters}

\vspace{0.5em}
\noindent
\begin{minipage}{\textwidth}
\centering
\begin{tabularx}{\textwidth}{l c X}
\toprule
\textbf{Criterion} & \textbf{Rating} & \textbf{Evidence} \\
\midrule
\faCheckCircle Reliability & + & Achieve performance close to full fine-tuning \citep{houlsby2019parameterefficienttransferlearningnlp}. \\
\addlinespace
\faArrowRight Generalization & + & Adapters generalize through composition. By separating knowledge extraction from knowledge composition, adapters effectively leverage representations learned from multiple tasks \citep{pfeiffer2021adapterfusion}. \\
\addlinespace
\faStar Specificity & 0 & Mixed evidence; some preservation of base capabilities but task-dependent \citep{pfeiffer-etal-2020-mad}. \\
\addlinespace
\faThermometerHalf Compute Efficiency & + & Small trainable modules reduce training cost \citep{houlsby2019parameterefficienttransferlearningnlp}. \\
\addlinespace
\faDatabase Data Efficiency & $-$ & Standard adapters require substantial task-specific data and struggle in low-resource settings \citep{pmlr-v188-bansal22a}. \\
\addlinespace
\faLink Composability & + & AdapterFusion and related methods enable structured composition \citep{pfeiffer2021adapterfusion}. \\
\addlinespace
\faUser Usability & $-$ & Training infrastructure still required. \\
\addlinespace
\faUndo Reversibility & + & Adapter modules can be removed or swapped without affecting base model. \\
\bottomrule
\end{tabularx}
\captionof{table}{Evidentiary support for Adapter ratings in Table \ref{Tab:criteria}.}
\label{tab:evidence_adapters}
\end{minipage}

\clearpage
\subsubsection{Soft Prompt Tuning}

\vspace{0.5em}
\noindent
\begin{minipage}{\textwidth}
\centering
\begin{tabularx}{\textwidth}{l c X}
\toprule
\textbf{Criterion} & \textbf{Rating} & \textbf{Evidence} \\
\midrule
\faCheckCircle Reliability & + & Achieves performance comparable to full fine-tuning \citep{lester2021powerscaleparameterefficientprompt}. \\
\addlinespace
\faArrowRight Generalization & + & Conditioning frozen models with soft prompts provides robustness to domain transfer compared to full model tuning \citep{lester2021powerscaleparameterefficientprompt}. \\
\addlinespace
\faStar Specificity & 0 & Operates at a coarse-grained task level. Standard prompt tuning methods learn task-level prompts that lack precision for fine-grained behavioral control \citep{chen2024fine,dai2025dual}. \\
\addlinespace
\faThermometerHalf Compute Efficiency & + & Small number of trainable parameters. \\
\addlinespace
\faDatabase Data Efficiency & 0 & Mixed results across prompt-based methods. Prefix tuning outperforms fine-tuning in low-data settings \citep{li-liang-2021-prefix}, while prompt tuning requires substantial labeled data \citep{guo-etal-2022-improving}. \\
\addlinespace
\faLink Composability & + & Soft prompts can be trained independently and flexibly composed via attention mechanisms. ATTEMPT demonstrates modular composition through interpolation of pre-trained source prompts \citep{asai-etal-2022-attempt}. \\
\addlinespace
\faUser Usability & $-$ & Training infrastructure still required. Prompting exhibits brittleness as an adaptation method \citep{10.1162/tacl_a_00681}. \\
\addlinespace
\faUndo Reversibility & + & Learned soft prompts easily removed. \\
\bottomrule
\end{tabularx}
\captionof{table}{Evidentiary support for Soft Prompt Tuning ratings in Table \ref{Tab:criteria}.}
\label{tab:evidence_softprompt}
\end{minipage}

\clearpage
\subsubsection{LoRA}

\vspace{0.5em}
\noindent
\begin{minipage}{\textwidth}
\centering
\begin{tabularx}{\textwidth}{l c X}
\toprule
\textbf{Criterion} & \textbf{Rating} & \textbf{Evidence} \\
\midrule
\faCheckCircle Reliability & + & Matches or exceeds full fine-tuning on many tasks \citep{hu2021loralowrankadaptationlarge}. \\
\addlinespace
\faArrowRight Generalization & + & Demonstrates strong transfer learning and domain generalization capabilities \citep{hu2021loralowrankadaptationlarge}. \\
\addlinespace
\faStar Specificity & 0 & Performs task-level adaptation by injecting low-rank matrices into model layers rather than enabling fine-grained behavioral control \citep{hu2021loralowrankadaptationlarge}. \\
\addlinespace
\faThermometerHalf Compute Efficiency & + & QLoRA further reduces memory requirements \citep{dettmers2023qlora}. \\
\addlinespace
\faDatabase Data Efficiency & 0 & Parameter efficiency does not directly translate to data efficiency. Some variants demonstrate improved few-shot learning \citep{chavan2023one}, but standard LoRA requires comparable training data to full fine-tuning \citep{hu2021loralowrankadaptationlarge}. \\
\addlinespace
\faLink Composability & + & Multiple independently trained LoRA modules can be combined to create multi-task models without additional training \citep{zhao2025merging,feng-etal-2024-mixture}. \\
\addlinespace
\faUser Usability & $-$ & Requires training infrastructure and hyperparameter tuning. While more accessible than full fine-tuning, still demands computational setup \citep{hu2021loralowrankadaptationlarge}. \\
\addlinespace
\faUndo Reversibility & + & LoRA interventions are fully reversible. \\
\bottomrule
\end{tabularx}
\captionof{table}{Evidentiary support for LoRA ratings in Table \ref{Tab:criteria}.}
\label{tab:evidence_lora}
\end{minipage}

\clearpage
\subsection{Steering Methods}

\subsubsection{Difference-based}

\vspace{0.5em}
\noindent
\begin{minipage}{\textwidth}
\centering
\begin{tabularx}{\textwidth}{l c X}
\toprule
\textbf{Criterion} & \textbf{Rating} & \textbf{Evidence} \\
\midrule
\faCheckCircle Reliability & + & Steering exhibits reliable in-distribution performance when tested on data distributions matching vector construction \citep{alex2023steering, rimsky-etal-2024-steering}. CAA demonstrates stable steering across test examples by averaging activation differences over multiple contrastive pairs to reduce variance \citep{rimsky-etal-2024-steering}. \\
\addlinespace
\faArrowRight Generalization & 0 & Cross-prompt and cross-domain transfer has been demonstrated \citep{arditi2024refusal}, but generalization to out-of-distribution prompts shows mixed results depending on the concept \citep{tan2024analysing}. Steering vectors often generalize well but can be brittle to prompt changes in some cases \citep{tan2024analysing}. \\
\addlinespace
\faStar Specificity & + & Steering preserves base model performance on unrelated tasks \citep{panickssery2023steering}. CAA minimally reduces general capabilities, with MMLU scores showing only 2-4\% degradation when applying steering vectors \citep{rimsky-etal-2024-steering}. \\
\addlinespace
\faThermometerHalf Compute Efficiency & + & Requires only a single forward pass to compute steering vectors with minimal inference overhead \citep{alex2023steering}. ActAdd introduces negligible computational cost compared to baseline inference \citep{alex2023steering}. \\
\addlinespace
\faDatabase Data Efficiency & + & Effective with few contrastive examples \citep{arditi2024refusal}. ActAdd can operate with as few as 2 prompt pairs \citep{alex2023steering}, while CAA typically uses dozens to hundreds of contrastive pairs for robust steering \citep{rimsky-etal-2024-steering}. Sample efficiency studies indicate that approximately 80-100 contrastive pairs per property are needed to avoid variance, with performance plateauing thereafter \citep{tan2024analysing}. \\
\addlinespace
\faLink Composability & 0 & Additive composition is possible and multiple steering vectors can be combined for multidimensional control \citep{ilharco2022editing}, but interference between non-orthogonal directions limits reliability. Feature composability is demonstrably robust when underlying concept vectors remain orthogonal in activation space \citep{tan2024analysing}. \\
\addlinespace
\faUser Usability & 0 & Requires identifying appropriate steering directions and intervention layers \citep{rimsky-etal-2024-steering}. The scaling coefficient for steering vectors requires empirical tuning, with acceptable ranges often being narrow \citep{alex2023steering}. \\
\addlinespace
\faUndo Reversibility & + & Interventions applied at inference; no parameter modification. \\
\bottomrule
\end{tabularx}
\captionof{table}{Evidentiary support for Difference-based Steering ratings in Table \ref{Tab:criteria}.}
\label{tab:evidence_difference}
\end{minipage}

\clearpage
\subsubsection{Optimization-based}

\vspace{0.5em}
\noindent
\begin{minipage}{\textwidth}
\centering
\begin{tabularx}{\textwidth}{l c X}
\toprule
\textbf{Criterion} & \textbf{Rating} & \textbf{Evidence} \\
\midrule
\faCheckCircle Reliability & + & LoReFT achieves state-of-the-art performance on commonsense reasoning and instruction-following tasks, demonstrating stable and consistent results \citep{wu2024reft}. Performance remains robust across different domains and model sizes. \\
\addlinespace
\faArrowRight Generalization & + & Task-level generalization demonstrated across diverse benchmarks including commonsense reasoning, arithmetic reasoning, instruction-following, and natural language understanding \citep{wu2024reft}. ReFT interventions transfer effectively across related tasks. \\
\addlinespace
\faStar Specificity & + & Targeted layer and position interventions allow precise control. ReFT selects specific timesteps and representations to intervene on, providing fine-grained behavioral modification \citep{wu2024reft}. \\
\addlinespace
\faThermometerHalf Compute Efficiency & 0 & Training phase required with gradient-based optimization, comparable to PEFT methods in computational cost \citep{wu2024reft}. However, ReFT is 15$\times$--65$\times$ more parameter-efficient than LoRA, requiring fewer trainable parameters. \\
\addlinespace
\faDatabase Data Efficiency & 0 & Requires training data comparable to PEFT methods \citep{wu2024reft}. No evidence of superior few-shot performance compared to other parameter-efficient approaches. \\
\addlinespace
\faLink Composability & 0 & Composition properties remain under-explored in the literature. While ReFT interventions can be defined independently, systematic evaluation of combining multiple ReFT modules is limited. \\
\addlinespace
\faUser Usability & 0 & New methodology with developing tooling. Requires PyReFT library and understanding of intervention design \citep{wu2024reft}. More accessible than full fine-tuning but requires expertise in selecting intervention layers and positions. \\
\addlinespace
\faUndo Reversibility & + & Interventions applied at inference time and can be removed without affecting the frozen base model \citep{wu2024reft}. Multiple task-specific ReFT interventions can be swapped dynamically. \\
\bottomrule
\end{tabularx}
\captionof{table}{Evidentiary support for Optimization-based Steering ratings in Table \ref{Tab:criteria}.}
\label{tab:evidence_optimization}
\end{minipage}

\clearpage
\subsubsection{Dictionary-based}

\vspace{0.5em}
\noindent
\begin{minipage}{\textwidth}
\centering
\begin{tabularx}{\textwidth}{l c X}
\toprule
\textbf{Criterion} & \textbf{Rating} & \textbf{Evidence} \\
\midrule
\faCheckCircle Reliability & 0 & Mixed results for steering effectiveness. While SAEs extract interpretable features \citep{cunningham2023sparse}, their steering performance is not competitive with simpler baselines \citep{wuaxbench}. Steering via SAE features can improve performance in some tasks but shows limitations compared to prompting and fine-tuning. \\
\addlinespace
\faArrowRight Generalization & + & Feature-level transfer demonstrated across different contexts \citep{deng-etal-2025-unveiling}. SAE features show language-specific and cross-lingual patterns, enabling transfer across related domains. Individual features can activate at multiple layers for different prompts. \\
\addlinespace
\faStar Specificity & + & Enables fine-grained, feature-level control. SAEs decompose dense activations into monosemantic features corresponding to specific semantic concepts \citep{cunningham2023sparse, gao2024scaling}. Individual features can be selectively manipulated for targeted behavioral modification. \\
\addlinespace
\faThermometerHalf Compute Efficiency & $-$ & Training SAEs is computationally expensive and data-intensive \citep{gao2024scaling}. Requires large activation corpora (billions of datapoints) and overcomplete dictionary representations. \\
\addlinespace
\faDatabase Data Efficiency & $-$ & Requires extensive activation data for training. SAEs typically need large-scale activation corpora to learn comprehensive feature dictionaries \citep{gao2024scaling}. Training datasets of 8 billion datapoints or more are common for robust feature extraction. \\
\addlinespace
\faLink Composability & 0 & Multiple SAE features can be combined for multidimensional control. Feature compositionality is demonstrably robust when features remain semantically distinct \citep{bayat2025steering}. However, direct decomposition of steering vectors using SAEs faces limitations due to negative projections and distribution mismatch \citep{mayne2024can}. \\
\addlinespace
\faUser Usability & 0 & Pre-trained SAE dictionaries (e.g., Gemma Scope \citep{lieberum-etal-2024-gemma}, LLaMA Scope \cite{he2024llama}) and interpretable feature labels enable more intuitive feature selection compared to raw activation steering. However, selecting task-relevant features from thousands of learned directions remains non-trivial \citep{zhao-etal-2026-denoising}. \\
\addlinespace
\faUndo Reversibility & + & Features applied at inference time without parameter modification. SAE-based steering operates on activations dynamically, allowing features to be added or removed without retraining. \\
\bottomrule
\end{tabularx}
\captionof{table}{Evidentiary support for Dictionary-based Steering ratings in Table \ref{Tab:criteria}.}
\label{tab:evidence_dictionary}
\end{minipage}